%% file: main.tex
\documentclass[manuscript]{acmart}

\renewcommand\footnotetextcopyrightpermission[1]{} 
\usepackage{multirow}
\usepackage{footnote}

\AtBeginDocument{%
  }

\setcopyright{acmlicensed}
\copyrightyear{2018}
\acmYear{2018}
\acmDOI{XXXXXXX.XXXXXXX}
\acmConference[Conference acronym 'XX]{Make sure to enter the correct
  conference title from your rights confirmation email}{June 03--05,
  2018}{Woodstock, NY}
\acmISBN{978-1-4503-XXXX-X/18/06}

\begin{document}

\title{Road Rage Reasoning with Vision-language Models (VLMs): Task Definition and Evaluation Dataset}


\author{Yibing Weng}
\affiliation{%
  \department{School of Computer Science and Engineering}
  \institution{University of Electronic Science and Technology of China}
  \city{Chengdu}
  \country{China}}
\email{202411081538@std.uestc.edu.cn}

\author{Yu Gu}
\affiliation{%
  \department{School of Computer Science and Engineering}
  \institution{University of Electronic Science and Technology of China}
  \city{Chengdu}
  \country{China}}
\email{yugu.bruce@ieee.org}

\author{Fuji Ren}
\affiliation{%
  \department{School of Computer Science and Engineering}
  \institution{University of Electronic Science and Technology of China}
  \city{Chengdu}
  \country{China}}
\email{renfuji@uestc.edu.cn}









\begin{abstract}
Road rage, triggered by driving-related stimuli such as traffic congestion and aggressive driving, poses a significant threat to road safety. Previous research on road rage regulation has primarily focused on response suppression, lacking proactive prevention capabilities. With the advent of Vision-Language Models (VLMs), it has become possible to reason about trigger events visually and then engage in dialog-based comforting before drivers' anger escalates. To this end, we propose the road rage reasoning task, along with a finely annotated test dataset and evaluation metrics, to assess the capabilities of current mainstream VLMs in scene understanding, event recognition, and road rage reasoning. The results indicate that current VLMs exhibit significant shortcomings in scene understanding within the visual modality, as well as in comprehending the spatial relationships between objects in the textual modality. Improving VLMs' performance in these areas will greatly benefit downstream tasks like antecedent-focused road rage regulation. 

\end{abstract}



\maketitle

\section{Introduction}


Road rage is defined as intense behaviorally maladaptive anger triggered by driving-related stimuli like traffic jams and aggressive driving \cite{galovski2006road}. It often leads to impulsive behaviors including swearing, deliberate ramming, and in extreme cases, even shooting \cite{wells2002exploratory}, posing a significant threat to road safety. It has attracted attention from various fields. In transportation, researchers mainly focus on designing scales to analyze the relationship between age, personality, and impulsive behavior \cite{failde2023traffic, an2023adaptation, youssef2023driving, ozturk2024mad}. In psychology, the generation mechanisms \cite{bjureberg2021regulating} and regulation strategies \cite{qu2024mindfulness, love2024development} of road rage have received much attention. In computer science, research focuses more on driver anger detection \cite{zepf2020driver}.

\begin{figure}
  \includegraphics[width=\textwidth]{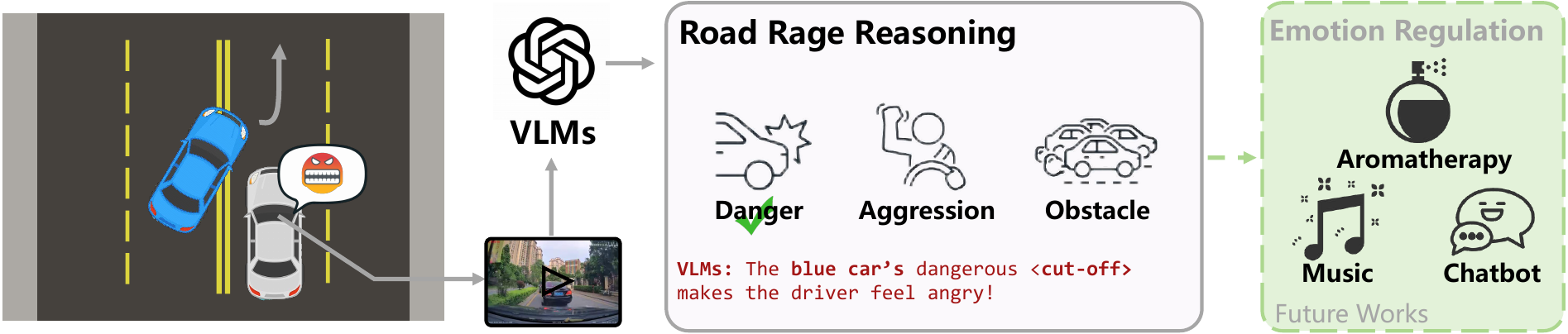}
  \caption{We propose the road rage reasoning task to evaluate VLMs' capabilities in road rage visual understanding, behavior recognition, and scenario reasoning. Provides prior knowledge for downstream emotion regulation tasks, enabling antecedent-focused regulation.}
  \label{fig:overview}
\end{figure}

As for the regulation of road rage, the focus of the aforementioned research lies on the anger emotion of drivers, and thus the resulting regulatory measures are primarily based on side-channels such as music \cite{fakhrhosseini2014if}, scent \cite{Dmitrijs2020CARoma}, and lighting \cite{hassib2019detecting} to suppress potential impulsive behaviors. In other words, these methods aim to calm drivers who are already in a state of anger, but lack the proactive regulation capability (e.g., antecedent-focused regulation) to intervene before anger forms. This situation is mainly caused by technical difficulties in two aspects: proactive regulation requires prompt and accurate understandings of the potential trigger events as well as natural interactive means (such as conversational soothing) to address these causes. With the recent success of VLMs in autonomous driving tasks (especially in understanding external driving environments) \cite{sima2025drivelm, ma2025dolphins, tian2024drivevlm}, it naturally raises the following question: \emph{\textbf{Can current VLMs handle these difficulties to enable antecedent-focused regulation (see Fig. ~\ref{fig:overview})?}}

As a first step in answering this question, we propose a road rage reasoning task for VLMs to evaluate how well they can understand the trigger events (i.e., antecedents) of potential road rage. The task takes visual inputs from a dashcam and assesses the capabilities of a VLM in scene understanding, event recognition, and road rage reasoning. The output is in text format, which can benefit downstream tasks like dialogue-based comforting, as shown in Fig. \ref{fig:task-definition}). 

To perform the task, we present a finely-annotated test dataset collected from real-world road rage dashcam footages along with various evaluation metrics. As shown in Fig. \ref{fig:data-labeling-overview}, the dataset consists of $81$ videos, $2,299$ frames, and $22,226$ annotations, which cover both overall labels for one footage (environment descriptions, road rage events, and road rage scenarios) and detailed labels for each frame (lane count, ego car, and critical objects). 


We tested two of the most widely-used VLMs, i.e., GPT-4o and Qwen-VL, on the task. The results revealed significant shortcomings in their visual scene understanding, which was crucial for event recognition. Furthermore, they struggled to comprehend spatial relationships between objects described in the text. For future work, we will try to fine-tune the VLMs with our dataset to enhance their capabilities in the above two aspects. 



\section{Related Works}

\subsection{Road Rage}
Currently, most studies on the causes of road rage use self-reports to understand which drivers are more prone to anger and which situations are more likely to trigger it \cite{abele2020links}. These insights can be used for modeling and automatically predicting when a driver may become angry. Some methods have also created simulated driving scenarios to assess whether road rage occurs in different situations \cite{wang2024inducing,lee2023effect}. These studies provide knowledge on the causes of road rage and suggest developing intervention strategies. Although some research proposes antecedent-focused regulation \cite{gross1998antecedent} to intervene before emotions occur, how to apply antecedent-focused regulation to road rage remains a challenge. This is because current techniques typically focus on detecting road rage after it has occurred, rather than identifying the causes that trigger it.

Common causes of road rage come from external driving environments \cite{deffenbacher2016review}. Although some driving datasets have been introduced \cite{caesar2020nuscenes,huang2018apolloscape,sun2020scalability}, which annotate objects in the environment and their potential threats to the ego car \cite{singh2022road,ettinger2021large}, these datasets mainly focus on training autonomous driving systems rather than perceiving external factors which cause road rage. There is still a lack of datasets specifically focused on road rage in driving environments.

\subsection{VLMs in Driving Tasks}
VLMs have made significant progress in the field of autonomous driving \cite{zhou2024vision}. By combining computer vision and natural language processing, VLMs can understand and process complex information encountered during driving, such as road signs \cite{cao2024maplm}, vehicle behavior, and pedestrian detection \cite{duan2024cityllava}. Numerous studies have shown that VLMs exhibit strong multimodal understanding abilities in driving environments, allowing them to comprehend both visual and textual inputs and make accurate inferences.

In recent studies (\cite{chen2024driving,sima2025drivelm,tian2024drivevlm}), VLMs have demonstrated impressive performance in driving and driving-related dialogue tasks, including visual understanding of the external environment, visual question answering based on that understanding, and the ability to infer the potential impact of external objects on the current vehicle. This shows the potential of VLMs to reason about the external driving environment that may trigger road rage.

\section{Task Definition and Test Dataset Collection}
\begin{figure}
    \centering
    \includegraphics[width=1\linewidth]{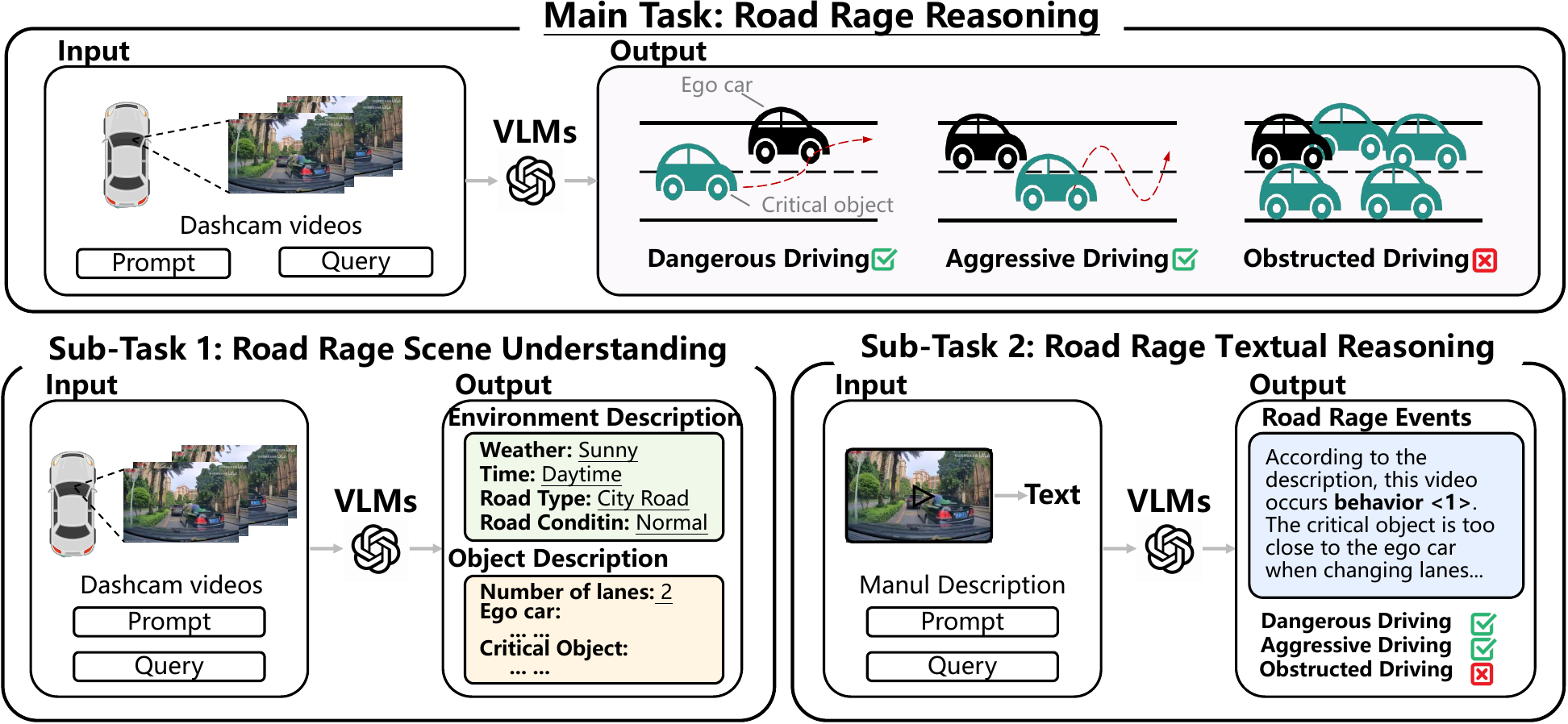}
    \caption{We design three tasks to evaluate VLMs' road rage reasoning abilities. The \textbf{Main Task} uses dashcam footage to identify road rage scenarios, testing overall reasoning. Due to poor performance, we introduce two subtasks. \textbf{Sub-task 1} uses dashcam footage to assess scene understanding but lacks complete responses for quantitative analysis. Thus, \textbf{Sub-task 2} uses manual descriptions, decoupling visual understanding from reasoning, and assesses textual reasoning and scene understanding capabilities.}
    \label{fig:task-definition}
\end{figure}

In this section, we define a main task and two sub-tasks to test VLMs' capabilities in road rage reasoning, (as shown in Fig. ~\ref{fig:task-definition}) and explain the rationale behind the design of these tasks. We then provided a detailed description of the dataset collection and annotation process, including the criteria for video collection and the content of the annotations.

\subsection{Task Definition}

\subsubsection{Main task: road rage reasoning}
The main task requires VLMs to understand the dashcam footage as a whole and determine whether dangerous, aggressive, or obstructive driving are present. This task implicitly demands full scene understanding and the ability to capture key features for situation assessment. Additionally, VLMs must recognize events in context and link them to corresponding road rage scenarios. Due to poor performance in the main task, we introduce two sub-tasks to explore the factors limiting VLMs' performance in road rage reasoning tasks.

\subsubsection{Sub-task1: road rage scene understanding}
Sub-task 1 also uses dashcam videos, where  VLMs are required to provide detailed descriptions of each frame based on a template we provide. The responses are in a fill-in-the-blank format, including environment and object descriptions (see Fig. ~\ref{fig:data-labeling-overview} (b)). The template defines options for each blank to aid subsequent calculations. Unfortunately, despite testing various prompts, VLMs still fail to describe each frame, preventing us from drawing quantifiable conclusions. Therefore, we set up sub-task 2.

\subsubsection{Sub-task2: road rage textual reasoning} 
Sub-task 2 converts the video labels into manual descriptions and uses them as input for VLMs. VLMs are then asked to recognize road rage events. Based on the output events of the VLMs, we categorize the road rage scenarios. VLMs can extract all relevant information related to road rage scenarios from manual descriptions. Therefore, by using this method, we can assume that VLMs have perfect scene understanding, decoupling scene understanding from reasoning. The conclusions from the experiment directly reflect VLMs' performance in textual reasoning and indirectly highlight their shortcomings in scene understanding.

\subsection{Dataset Collection}

\input{samples/images/data-labeling-overview}

Fig. ~\ref{fig:data-labeling-overview} is an overview of the dataset. It shows the number of videos we collected, the number of frames divided, the total number of annotations, and provides an example of video annotation.

\subsubsection{Selection criteria}
To create the road rage reasoning dataset, we recruit three volunteers, each with at least two years of driving experience. They manually select videos from popular platforms such as YouTube, Bilibili, and Youku. The main criterion for video selection is: "Does the video make you feel angry, anxious, or tense, and does it show at least one of the following: dangerous driving, aggressive driving, or obstructive driving?". Additionally, to standardize the content, the following restrictions are applied:

\begin{itemize}
    \item The video must be recorded from the driver’s first-person perspective.
    \item The video should not contain commentary subtitles.
    \item The events in the video must be easily observable and describable.
    \item The video must cover the entire duration of the event.
    \item The videos must be from the same region, as driving rules may vary by location.
\end{itemize}

Through this process, over 100 driving videos are collected, each initially annotated with a road rage scenario classification label.

\subsubsection{Road rage events labeling}
In our approach, we focus on having VLMs recognize the specific events that provoke anger, rather than just classifying scenarios into general categories. To achieve this, we annotate the road rage events in the collected videos. To ensure the events in the filtered videos are sufficiently anger-inducing, we refer to the MAD \cite{stephens2019measure}. The MAD scale assesses the level of anger triggered by different driving events. Based on the anger scores from MAD, we recalculate rankings using a weighted average method. 

Additionally, since some events in the MAD scale conflict with the rules for the videos we collected, we replace these events accordingly. After this, we apply secondary filtering to the videos based on two criteria: first, the anger score induced by the event should be as high as possible, and second, each event should have at least 10 samples.

\subsubsection{Scene labeling}
The next step involves detailed annotation of the road conditions, the behaviors of the ego car, and the behaviors of the critical objects involved in the road rage scenario. For the entire video, we first provide environment descriptions, including weather, time, road type, and road conditions. Then, we sample the video at a rate of two frames per second. For each sampled frame, we perform object descriptions, including the number of lanes on the current road, as well as the positions and behaviors of the ego car and critical objects. For the ego car, we describe its lane, the types of lane markings on both sides of the lane, and its actions (the actions are based on the DriveVLM  \cite{tian2024drivevlm}). For the critical object, we first describe its visual features, followed by its lane, its relative position to the ego car, the distance between them, and its actions. To facilitate subsequent calculations, we set predefined options for each variable. Fig. ~\ref{fig:data-labeling-overview} (b) shows an example of our annotation.

\subsubsection{Dataset statistics}

\input{samples/tables/type-num}

After filtering, we ultimately retain 81 videos. Based on the time range in which road rage events occur, and after sampling two frames per second, we obtain a total of 2,299 frames. We perform both overall annotations for the videos and detailed per-frame annotations, resulting in a total of 22,226 annotations. Fig. ~\ref{fig:data-labeling-overview} (a) provides a reference. In Table ~\ref{tab:type-num}, we list all the road rage events that occur in the videos, along with their frequency (note that multiple events can appear in a single video). These events correspond to specific road rage scenarios. Among them, dangerous driving includes 6 types of behaviors: 5 related to dangerous or illegal actions of vehicles, and 1 related to non-vehicle objects crossing the road. Aggressive driving consists of 2 types of behaviors: repeatedly braking to block the ego car and repeatedly cutting into the ego car's lane. Obstructive driving includes 1 situation: road congestion or heavy traffic.

\section{Evaluation and Results}
We select gpt-4o-2024-08-06 and qwen-vl-max-2024-11-19 as baseline models. All experiments are conducted in a zero-shot setting. The evaluation metrics include precision, recall, and accuracy, which help us assess VLMs' misjudgment rate, their ability to correctly identify specific scenarios or events, and their overall reasoning capability.

\input{samples/tables/scenario-classification}

\subsection{Evaluation of Main Task}
Firstly, we present the results of the main task, as shown in Table ~\ref{tab:main-task-result}. These results reflect VLMs' ability in overall reasoning. Based on the data from the table, we summarize our findings as follows:

\textbf{VLMs show significant room for improvement in reasoning road rage scenarios:} We observe that both GPT-4o and Qwen-VL show low accuracy in reasoning road rage based on the whole video, indicating that VLMs can't effectively understand video content.

\textbf{VLMs exhibit significant performance differences across different road rage scenarios:} The recall rates show that Qwen-VL performs well for visually-oriented obstructive driving, while GPT-4o performs slightly worse. For dangerous driving, VLMs can partially classify the scenarios. However, for reasoning-oriented aggressive driving, VLMs show the worst recall rates. This makes sense because obstructive driving is easier to detect due to the presence of many surrounding vehicles. Most dangerous driving incidents in our dataset involve critical object being very close to ego car, which makes it easier for VLMs to identify if VLMs can capture this feature. Aggressive driving, however, requires recognizing the events first and then inferring its aggressive nature, which is a significant challenge for current VLMs.

\textbf{VLMs tend to make misjudgments:} This conclusion is derived from comparing precision values. For dangerous driving, while recall rates show that not all dangerous behaviors are recognized, the precision is high. This indicates that VLMs make some correct judgments. For obstructive driving, while VLMs demonstrate relatively good recall, precision shows that there are still many misjudgments. For aggressive driving, precision is still poor.

\subsection{Evaluation of Sub-task 1}
\begin{figure}
    \centering
    \includegraphics[width=1\linewidth]{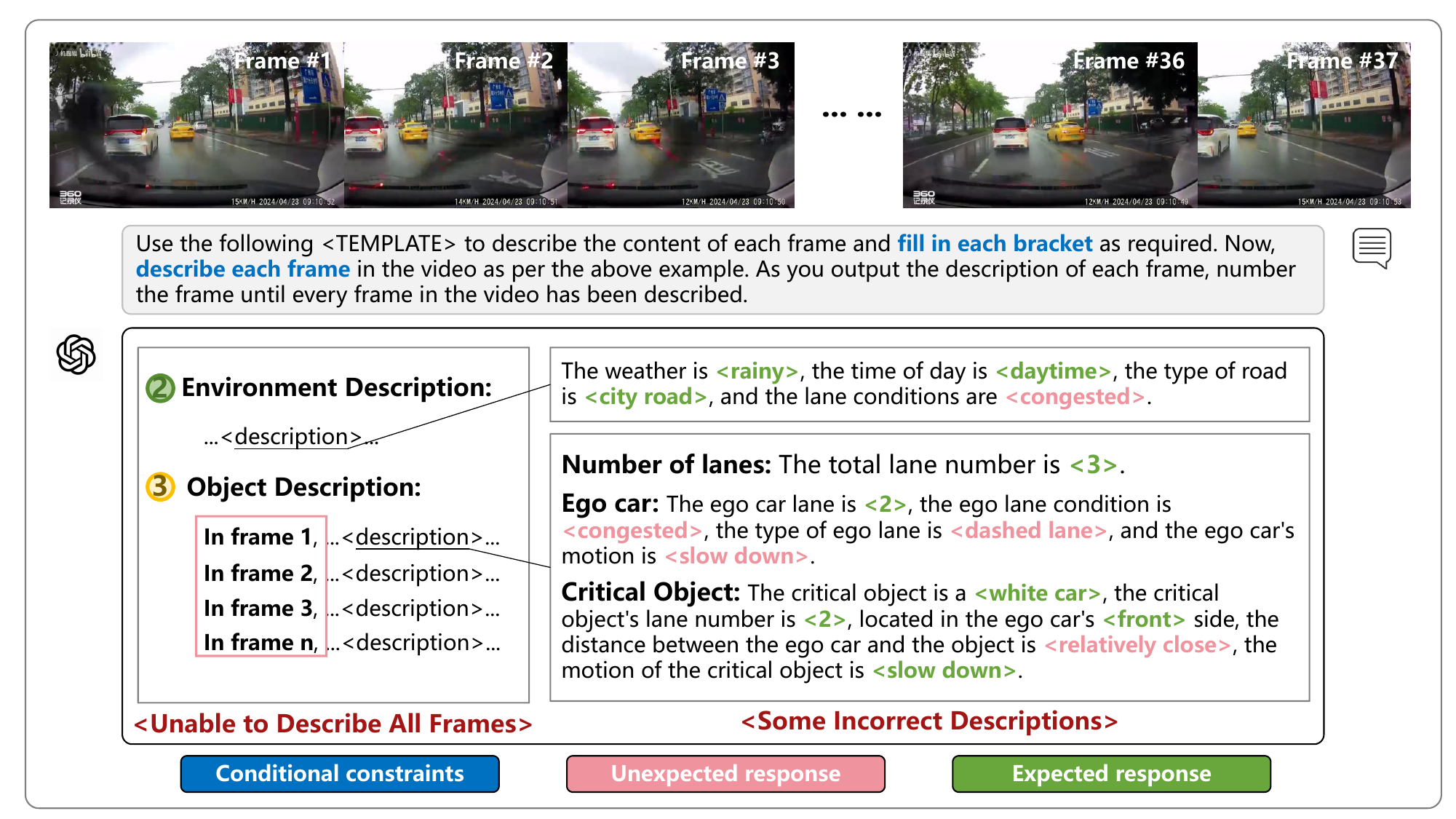}
    \caption{An experimental result from sub-task 1. Under the given constraints, VLMs still fail to describe all frames. In the details, VLMs show some incorrect descriptions. This result prevents us from performing a quantitative analysis of VLMs' visual understanding ability. Therefore, we introduce sub-task 2.}
    \label{fig:sub-task1-result}
\end{figure}

Sub-task 1 is a scene understanding task. In the prompt, we provide constraints and ask VLMs to fill in the blanks based on the template. VLMs must also describe each frame of the input. Fig. ~\ref{fig:sub-task1-result} shows an example answer for sub-task 1. 

The answers consist of two parts: environment description and object description (see Section 3.2). From the object description, we see that, despite the constraints, VLMs still cannot describe every frame. Additionally, there are some incorrect descriptions in both environment and object details. These results show that VLMs have limitations in scene understanding. Since VLMs cannot fully answer the question, we also cannot perform a quantitative analysis or draw firm conclusions. Therefore, we conduct additional experiments (Sub-task 2) to further explore VLMs' understanding and reasoning abilities.

\subsection{Evaluation of Sub-task 2}

\input{samples/tables/textual-reasoning-9-type}
\input{samples/tables/textual-reasoning-3-type}

Sub-task 2 converts video inputs into manual descriptions, assuming VLMs have perfect scene understanding, thereby decoupling understanding from reasoning. Table ~\ref{tab:textual-reasoning-9-type} presents the results of recognition of road rage events, and Table ~\ref{tab:textual-reasoning-3-type} shows the classification outcomes of road rage scenarios from these events. We summarize our findings as follows:

\textbf{VLMs' scene understanding limits road rage reasoning:} Table ~\ref{tab:textual-reasoning-9-type} demonstrates that VLMs recognize most events accurately and achieve high precision and recall when using manual descriptions as input. Table ~\ref{tab:textual-reasoning-3-type} compares road rage scenario classification results using video and textual modalities. Except for a slight drop in precision for some scenarios, other metrics show significant improvement. This highlights that limitations in scene understanding restrict VLMs' ability to reason road rage events and scenarios.

\textbf{VLMs’ spatial relationship comprehension limits textual reasoning:} Table ~\ref{tab:textual-reasoning-9-type} reveals that precision for aggressive driving events (Events 7 and 8) remains low. Additionally, the recall rate for Event 3 (critical objects crossing solid lines to overtake) has room for improvement. These results indicate that, even with complete scene information, VLMs struggle with tasks requiring spatial relationship comprehension, particularly for Events 3, 7, and 8.

\textbf{VLMs still perform misjudgment:} Despite significant improvements in road rage events recognition and scenarios classification when converting the visual modality to the manual description, VLMs still perform event misjudgments. Specifically, the precision for certain events, particularly aggressive driving events, remains low. This indicates that many events are mistakenly classified as aggressive.

\section{Discussion and Future Works}

\subsection{Discussion}
Based on the above experiments, we summarize two key observations: 1. VLMs’ limitations in scene understanding for road rage reasoning. While VLMs have made progress in identifying visual features, their ability to understand and reason about complex events remains a challenge. 2. VLMs’ shortcomings in spatial relationship comprehension with textual modality. Converting video inputs into textual modality assumes perfect scene understanding, yet VLMs struggle with spatial relationship comprehension tasks. Then, we propose two open questions:

1. How can VLMs’ understanding of complex scenes be enhanced?
Despite advances in visual feature recognition, enabling VLMs to understand and reason more intricate events remains a challenge. Could stronger reasoning modules or optimized training and fine-tuning strategies address this issue in future research?

2. How can VLMs’ spatial relationship comprehension in the text modality be improved?
Even with perfect scene understanding through textual inputs, VLMs exhibit significant weaknesses in spatial relationship comprehension, particularly in tasks that involve cross-frame comprehension of vehicle positions and movements. How can we enhance VLMs’ capabilities in spatial relationship comprehending within the text modality, especially for tasks requiring multi-frame understanding?

\subsection{Future Works}
Based on the two questions raised above, we discuss feasible solutions as future work:

\textbf{1. Fine-tuning VLMs using our collected dataset or existing datasets.} Existing studies and our experimental results indicate that current VLMs perform poorly in specific tasks like driving. The likely reasons are a lack of driving-related content in the training data and insufficient understanding of movements. Therefore, future work could involve fine-tuning VLMs using driving datasets (including image and action annotations) through instruction-based fine-tuning or full parameter fine-tuning. Given the lack of action annotations in existing driving datasets, a potential solution might be to fine-tune action understanding datasets first, then transfer to driving tasks. Alternatively, deep learning-based methods, which excel in object recognition and action understanding, could be explored.

\textbf{2. Fine-tuning VLMs for spatial relationship understanding in textual modality.} Most VLMs use the CLIP architecture, which connects visuals to text. Improving VLMs' understanding of spatial relationships in text can also boost their visual reasoning. To do this, future work can add special training for spatial relationships in textual reasoning. Using data that focuses on spatial relationships (e.g., object positions, speed changes) can guide VLMs to learn how to infer spatial changes in different situations. Reinforcement learning methods could be used to improve the model's ability to understand spatial relationships across frames.

\section{Conclusion}
We propose the road rage reasoning tasks and evaluate VLMs' scene understanding and textual reasoning abilities. We set up three tasks around road rage reasoning. The main task tests VLMs' overall reasoning ability, but the performance is poor. Sub-task 1 tests VLMs' scene understanding, but they fail to provide complete answers. Sub-task 2 converts the video into manual descriptions, decoupling visual understanding from reasoning. The results show that VLMs have limitations in scene understanding and spatial relationship comprehension in road rage reasoning. Therefore, we raise open questions, hoping to solve these two issues. Solving these problems can provide prior knowledge for downstream tasks like emotion regulation, contributing to a safer and more comfortable driving environment.

\bibliographystyle{ACM-Reference-Format}
\bibliography{main}

\appendix

\section{Dataset Details}
To facilitate the analysis of VLMs' responses, we define a set of possible options for each labeled variable. VLMs can only choose from these options, as shown in Table ~\ref{tab:data-labeling-details}. It’s important to note that the actions of critical objects are often complex and varied. To address this, we categorized the actions into general actions, vehicle actions, vehicle signals, and other actions. For vehicle objects, we use general actions, vehicle actions, and vehicle signals. For non-vehicle objects, we use general actions and other actions to describe their behavior.

\input{samples/tables/data-labeling-details}

\section{Experiment Results}
\begin{figure}[h]
    \centering
    \includegraphics[width=1\linewidth]{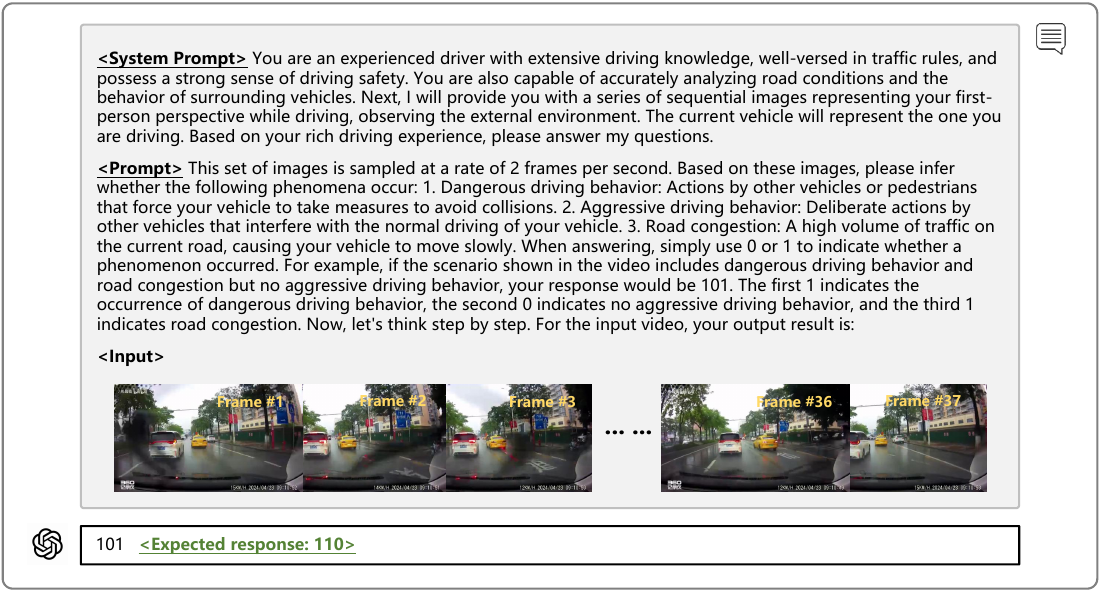}
    \caption{The experimental results for the main task are shown in the figure. We use video frames as input and ask the VLMs to identify dangerous driving, aggressive driving, and obstructive driving in the video. To simplify result analysis, VLMs are required to output a binary response (0 or 1).}
    \label{fig:main-task-exp}
\end{figure}

\begin{figure}
    \centering
    \includegraphics[width=0.85\linewidth]{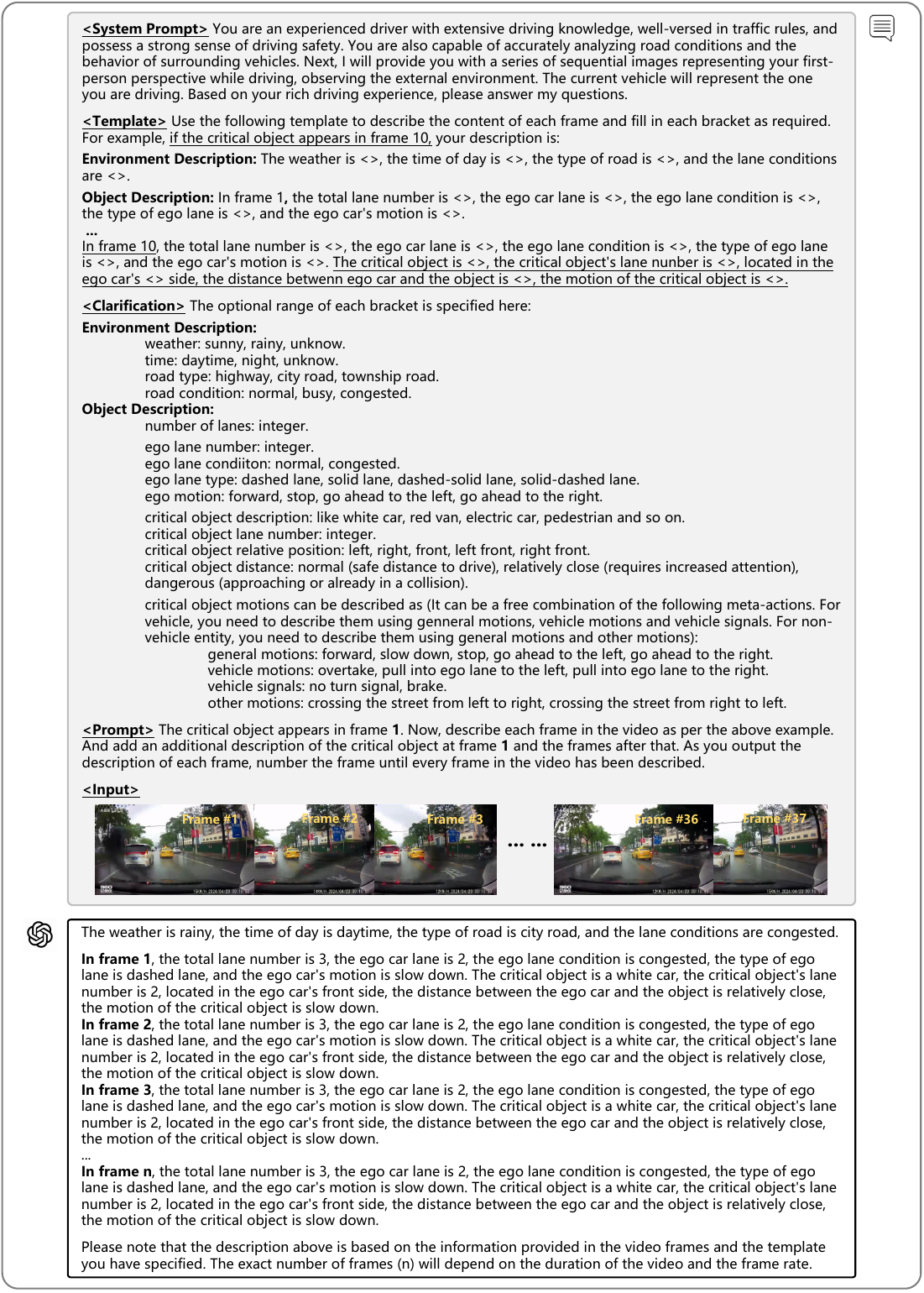}
    \caption{The experimental results for sub-task 1 are shown in the figure. We provide a response template and ask VLMs to describe each frame of the input video. Since the critical object does not appear in every frame of the video, we inform VLMs in advance about which frames contain the critical object. VLMs are required to describe the critical object in these frames and the following ones. The results show that VLMs fail to provide complete responses, preventing us from performing quantitative analysis.}
    \label{fig:sub-task1-exp}
\end{figure}

\begin{figure}
    \centering
    \includegraphics[width=0.9\linewidth]{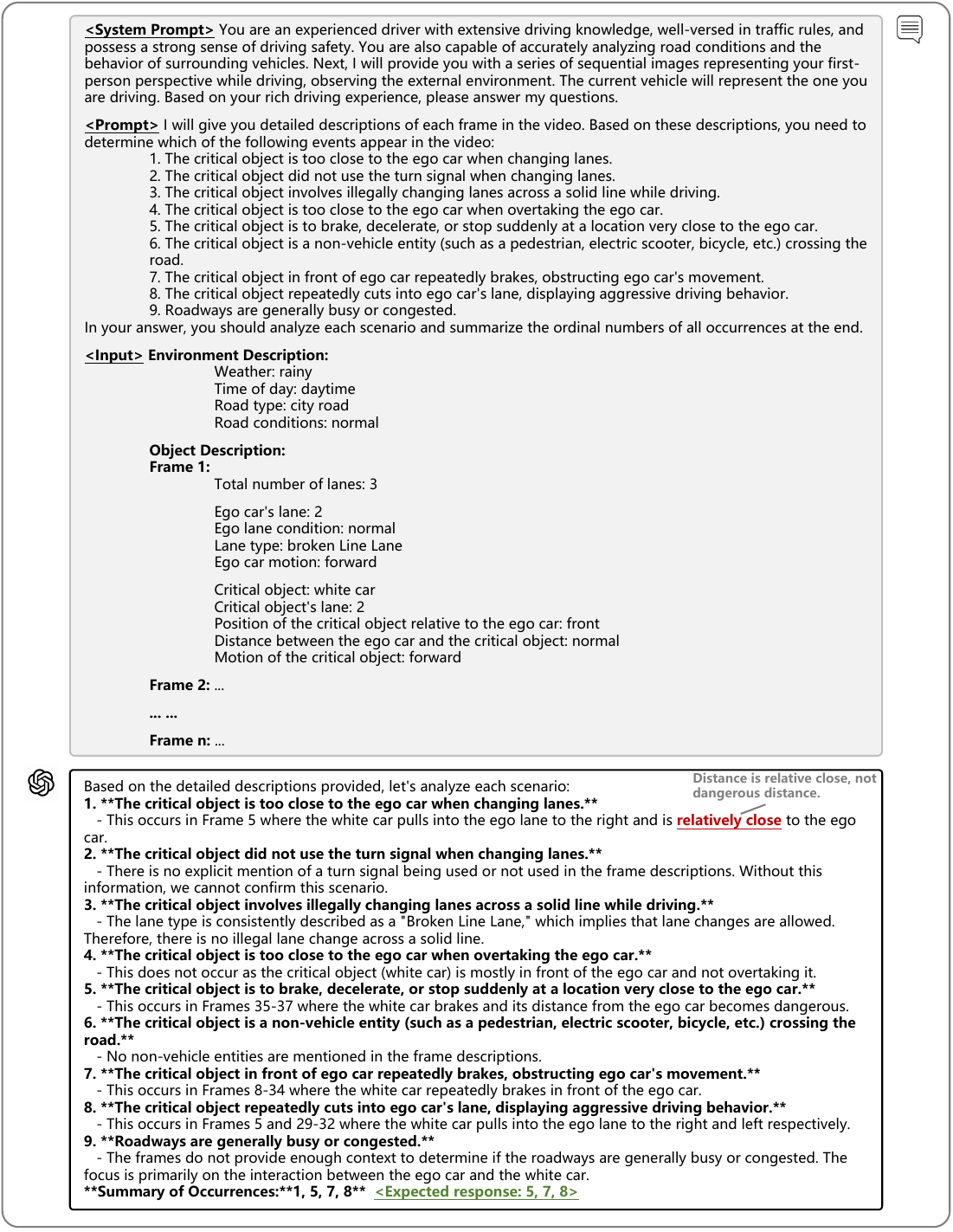}
    \caption{The experimental results for sub-task 2 are shown in the figure. We convert the labels of the video into manual descriptions and use these as input for VLMs. VLMs are tasked with reasoning which behaviors appear in the video based on the descriptions (the collected videos contain 9 types of behaviors). The results show that VLMs can reason most behaviors correctly, but there are some errors in the details.}
    \label{fig:sub-task2-exp}
\end{figure}

Fig. ~\ref{fig:main-task-exp}, ~\ref{fig:sub-task1-exp}, ~\ref{fig:sub-task2-exp} provide examples from the experiments of the main task, Sub-task 1, and Sub-task 2. These include how we set the prompts, the input formats, the output constraints, and the complete responses from the VLMs. 

\begin{figure}
    \centering
    \includegraphics[width=1\linewidth]{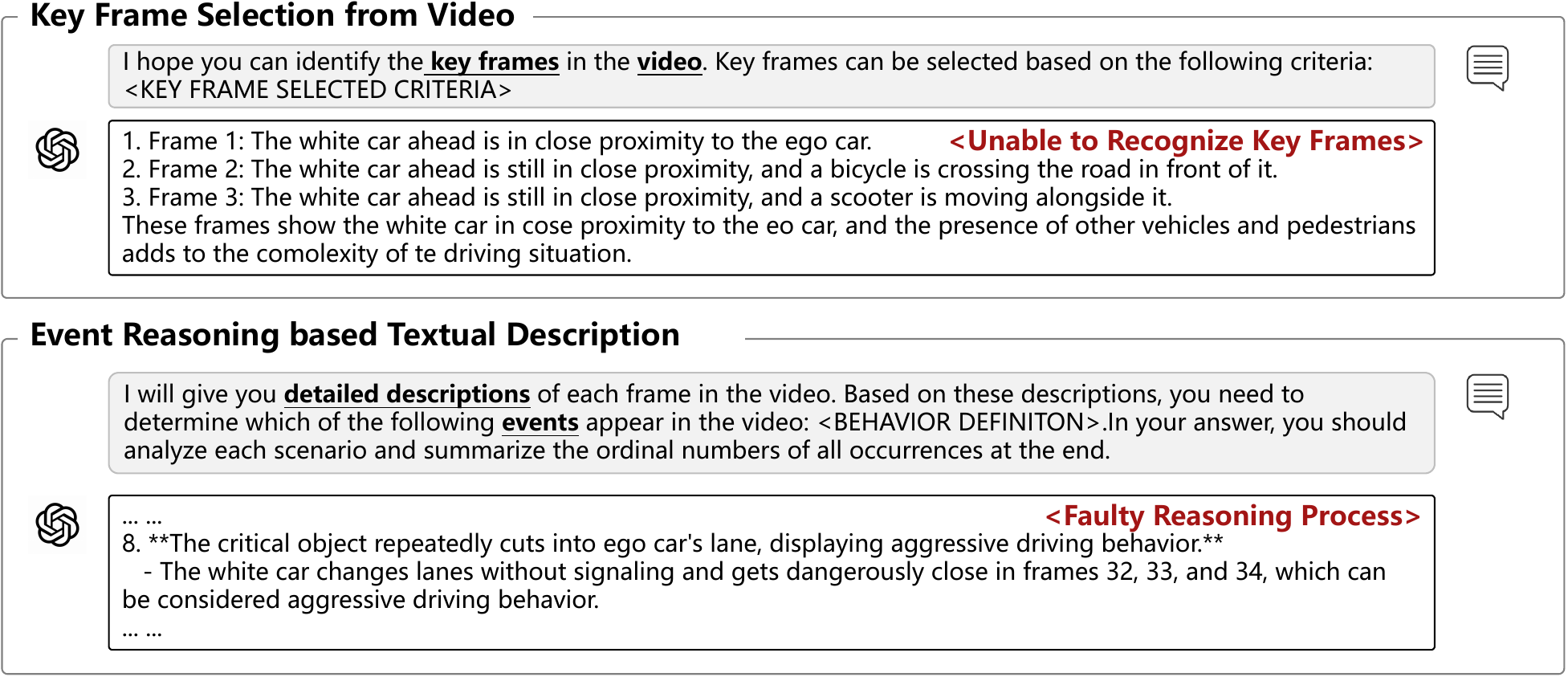}
    \caption{The results for selecting key frames from video input and performing event recognition based on textual description. The conclusion is that VLMs cannot identify key frames as requested, and they struggle with understanding spatial relationships in the textual modality to recognize road rage events.}
    \label{fig:observation}
\end{figure}

We also provide two experimental cases to explain our observations, as shown in Fig. ~\ref{fig:observation}. For scene understanding, we ask VLMs to select key frames based on prompts (e.g., when the critical object is too close to the ego vehicle). VLMs fail to judge distance visually and miss some key frames. For textual reasoning, we ask if event 8 (critical object repeatedly cuts into ego car’s lane) appears based on manual descriptions. VLMs fail to deduce the correct conclusion from spatial changes, showing irrelevant reasoning. This suggests that VLMs need to improve the understanding of spatial relationships in text.

\section{Exploration of Future Works}
\begin{figure}
    \centering
    \includegraphics[width=1\linewidth]{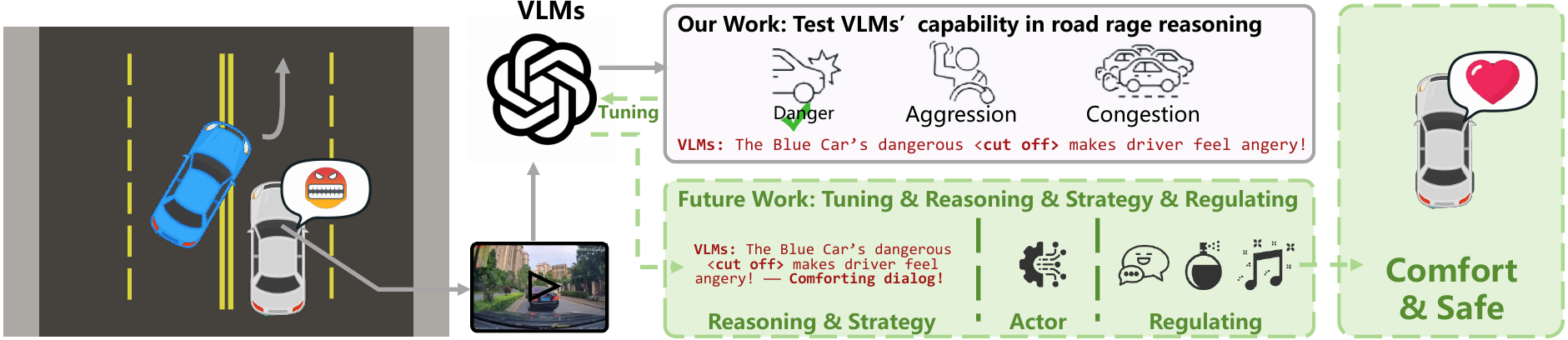}
    \caption{Future works}
    \label{fig:future-works}
\end{figure}

Our future work will first focus on fine-tuning VLMs to enable them to perceive external driving environments and reason about road rage. Then, we will explore emotional regulation methods, such as comforting conversations, aromatherapy, and music-based interventions. As shown in Fig. ~\ref{fig:future-works}. The ideal scenario is for VLMs to predict potential causes of road rage before it occurs, and to implement targeted strategies using emotional regulation techniques to alleviate the driver's road rage. This work could contribute to making future driving environments safer and more comfortable. 

\end{document}

%% file: samples/images/data-labeling-overview.tex
\begin{figure}
    \begin{minipage}[]{0.25\textwidth}
        \centering
        \resizebox{1.05\linewidth}{!}{
        \begin{tabular}{ll}
        \hline
        \textbf{Video selection}          &                        \\ \hline
        Main source                       & www.bilibili.com               \\
        Annotation language               & en / ch                \\
        Number of videos                      & 81                     \\
        Sampling rate                     & 2 fps                   \\
        Total frames                      & 2299                   \\
        Resolution                        & 512 $\times$ 288 pixels               \\
        Number of frames                  & 15 - 37                \\ \hline
        \textbf{Details}                  &                        \\ \hline
        Weather                           & \multicolumn{1}{l}{3}  \\
        Road types                        & \multicolumn{1}{l}{3}  \\
        Road conditions                    & \multicolumn{1}{l}{3}  \\
        Ego car's actions                 & \multicolumn{1}{l}{4}  \\
        Critical object's actions         & \multicolumn{1}{l}{10} \\ \hline
        \textbf{Annotations}              &                        \\ \hline
        Road rage scenarios  & 81 $\times$ 3                   \\
        Road rage events               & 81 $\times$ 9                   \\
        Environment descriptions                 & 81 $\times$ 4                   \\
        Number of lanes                       & 2299 $\times$ 1                 \\
        Ego car annotations                & 2299 $\times$ 4                 \\
        Critical object annotations        & 1887 $\times$ 5                 \\ \hline
        \textbf{Total}                    & 22226                  \\ \hline
        \end{tabular}%
        }
        \footnotetext{(a) Dataset statistic}
    \end{minipage}
    \begin{minipage}[]{0.7\textwidth}
        \centering
        \includegraphics[width=0.94\linewidth]{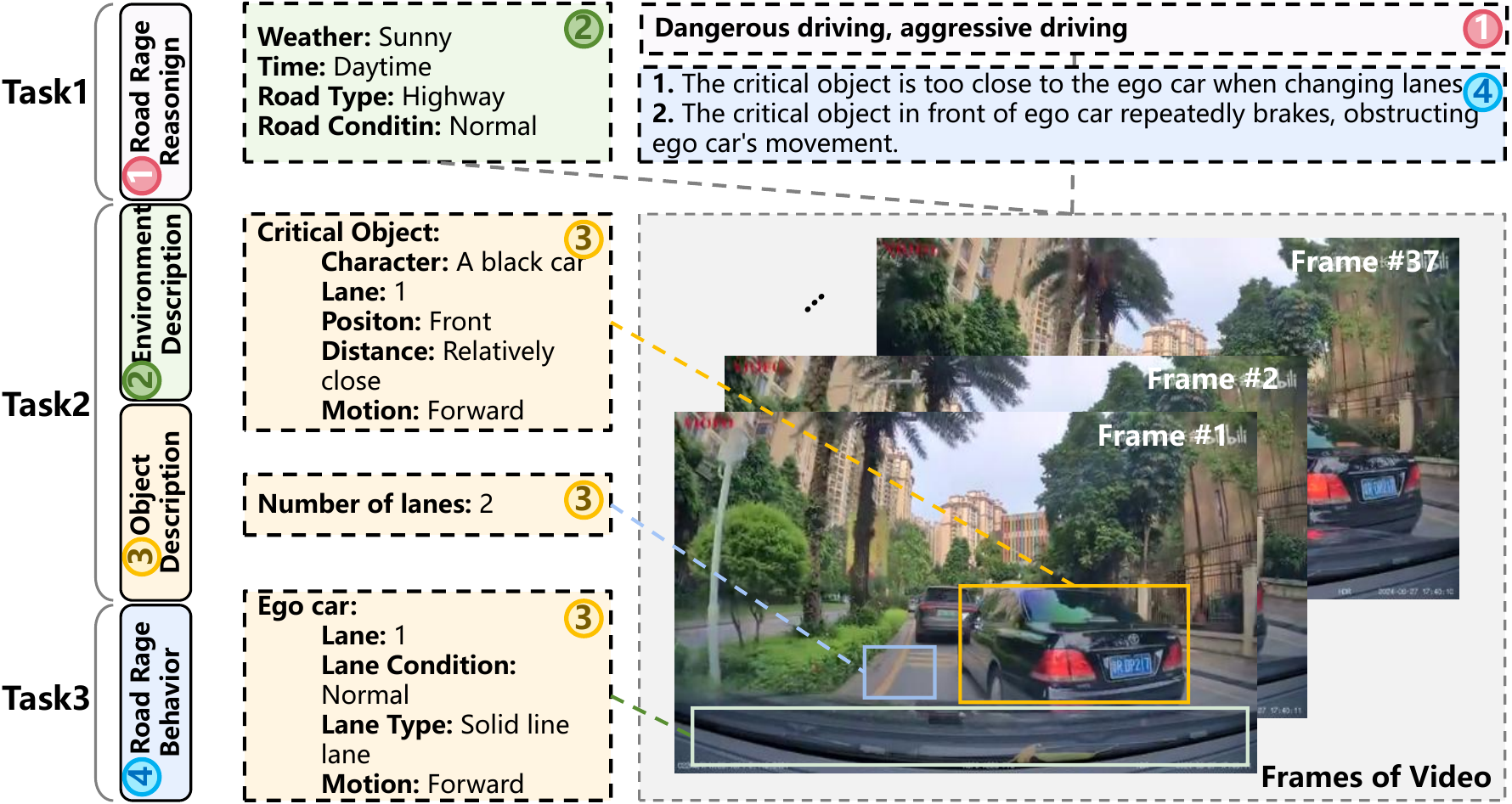}
        \footnotetext{(b) Dataset labeling sample}
    \end{minipage}
    \caption{The statistics (a) and an annotation example (b) of our dataset. The dataset includes 81 videos, 2,299 frames, and 22,226 annotations. The annotations cover both overall labels (environment descriptions, road rage events and road rage scenarios) and detailed labels (lane count, ego car, and critical objects).}
    \label{fig:data-labeling-overview}
\end{figure}

%% file: samples/tables/type-num.tex
\begin{table}
\caption{The number of specific events in the collected videos of the three road rage scenarios.}
\label{tab:type-num}
\resizebox{\textwidth}{!}{%
\begin{tabular}{llc}
\toprule
Type                                & Event Description                                                                                                   & Number \\ \midrule
\multirow{6}{*}{Dangerous Driving}  & 1. The critical object is too close to the ego car when changing lanes.                                                   & 44              \\
                                    & 2. The critical object did not use the turn signal when changing lanes.                                                   & 27              \\
                                    & 3. The critical object involves illegally changing lanes across a solid line while driving.                               & 14              \\
                                    & 4. The critical object is too close to the ego car when overtaking the ego car.                                           & 16              \\
                                    & 5. The critical object is to brake, decelerate, or stop suddenly at a location very close to the ego car.                 & 39              \\
                                    & 6. The critical object is a non-vehicle entity (such as a pedestrian, electric scooter, bicycle, etc.) crossing the road. & 11              \\ \hline
\multirow{2}{*}{Aggressive Driving} & 7. The critical object in front of ego car repeatedly brakes, obstructing ego car's movement.                             & 19              \\
                                    & 8. The critical object repeatedly cuts into ego car's lane, displaying aggressive driving behavior.                       & 13              \\ \hline
Obstructive Driving                  & 9. Roadways are generally busy or congested.                                                                              & 32              \\ \bottomrule
\end{tabular}%
}
\end{table}

%% file: samples/tables/scenario-classification.tex
\begin{table}
\centering
\caption{The performance of GPT-4o and Qwen-VL on the main task. Bold text and underlining represent the maximum and minimum values of precision and recall, respectively. VLMs perform best in obstructive driving, which is more visually-oriented, and worst in aggressive driving, which requires more reasoning ability.}
\label{tab:main-task-result}
\begin{tabular}{c|cc|cc|cc|c}
\hline
\multirow{2}{*}{Models} & \multicolumn{2}{c|}{Dangerous driving} & \multicolumn{2}{c|}{Aggressive driving} & \multicolumn{2}{c|}{Obstructed Driving} & \multirow{2}{*}{Accuracy} \\ \cline{2-7}
                        & Precision          & Recall            & Precision          & Recall             & Precision          & Recall             &                           \\ \hline
GPT-4o                  & \textbf{100.0}      & 31.17     & \underline{12.50}      & \underline{12.50}      & 61.11      & \textbf{68.75}      & 14.81             \\
Qwen-VL                 & \textbf{100.0}      & 28.57     & \underline{50.00}      &  \underline{4.17}       & 62.79      & \textbf{84.38}      & 18.52             \\ \hline
\end{tabular}
\end{table}

%% file: samples/tables/textual-reasoning-9-type.tex
\begin{table}
\centering
\caption{The experimental results of sub-task 2, where the underlined values represent the minimum. Using manual descriptions as input, VLMs can accurately reason most road rage events. However, there is still room for improvement in events requiring spatial relationship comprehension (Events 3, 7, and 8).}
\label{tab:textual-reasoning-9-type}
\resizebox{\textwidth}{!}{
\begin{tabular}{c|c|ccccccccc}
\hline
\multirow{2}{*}{Models}  & Road rage scenarios & \multicolumn{6}{c|}{Dangerous driving}                             & \multicolumn{2}{c|}{Aggressive driving} & Obstructive driving \\ \cline{2-11} 
                         & Road rage events & 1     & 2     & 3     & 4     & 5     & \multicolumn{1}{c|}{6}     & 7        & \multicolumn{1}{c|}{8}       & 9                  \\ \hline
\multirow{3}{*}{GPT-4o}  & Precision         & 87.76 & 81.82 & 77.78 & 88.89 & 90.70 & \multicolumn{1}{c|}{100.0} & \underline{45.24}    & \multicolumn{1}{c|}{58.82}   & 100.0              \\
                         & Recall            & 97.73 & 100.0 & \underline{50.00} & 100.0 & 100.0 & \multicolumn{1}{c|}{100.0} & 100.0    & \multicolumn{1}{c|}{76.92}   & 100.0              \\ \cline{2-11} 
                         & Accuracy          & \multicolumn{9}{c}{39.51}                                                                                                         \\ \hline
\multirow{3}{*}{Qwen-VL} & Precision         & 83.02 & 67.50 & 68.75 & 84.21 & 86.67 & \multicolumn{1}{c|}{100.0} & 46.15    & \multicolumn{1}{c|}{\underline{31.58}}   & 100.0              \\
                         & Recall            & 100.0 & 100.0 & \underline{78.57} & 100.0 & 100.0 & \multicolumn{1}{c|}{100.0} & 94.74    & \multicolumn{1}{c|}{92.31}   & 96.88              \\ \cline{2-11} 
                         & Accuracy          & \multicolumn{9}{c}{29.63}                                                                                                         \\ \hline
\end{tabular}}
\end{table}

%% file: samples/tables/textual-reasoning-3-type.tex

\begin{table}
\centering
\caption{The road rage scenarios derived from the events output by sub-task 2, with the results from VLMs based on video input shown after the slash. Except for a slight decrease in precision for some scenarios, other metrics have shown significant improvement.}
\label{tab:textual-reasoning-3-type}
\resizebox{\textwidth}{!}{
\begin{tabular}{c|cc|cc|cc|c}
\hline
\multirow{2}{*}{Models} & \multicolumn{2}{c|}{Dangerous driving} & \multicolumn{2}{c|}{Aggressive driving} & \multicolumn{2}{c|}{Obstructed Driving} & \multirow{2}{*}{Accuracy} \\ \cline{2-7}
                        & Precision          & Recall             & Precision          & Recall              & Precision           & Recall              &                           \\ \hline
GPT-4o                  & 96.25 / 100.0       & 100.0 / 31.17 $\uparrow$     & 46.00 / 12.50 $\uparrow$     & 95.83 / 12.50 $\uparrow$     & 100.0 / 61.11 $\uparrow$      & 100.0 / 68.75 $\uparrow$     & 62.96 / 14.81 $\uparrow$            \\
Qwen-VL                 & 96.25 / 100.0      & 100.0 / 28.57 $\uparrow$       & 39.66 / 50.00       & 95.83 / 4.17 $\uparrow$      & 100.0 / 62.79 $\uparrow$     & 100.0 / 84.38 $\uparrow$     & 53.09 / 18.52 $\uparrow$            \\ \hline
\end{tabular}}
\end{table}

%% file: samples/tables/data-labeling-details.tex
\begin{table}[h]
\centering
\caption{The details of environment description and object description, including the selectable options for each variable.}
\resizebox{\textwidth}{!}{
\begin{tabular}{l|l|l|l} 
\hline
Descriptions                         & Variable                                  & \multicolumn{2}{l}{Options}                                                                                                                            \\ 
\hline
\multirow{4}{*}{Environment Description}   & weather                                  & \multicolumn{2}{l}{sunny/rainy/unknow}                                                                                                                 \\
                                     & time                                     & \multicolumn{2}{l}{daytime/night /unknow}                                                                                                              \\
                                     & road type                                & \multicolumn{2}{l}{highway/city road/township road}                                                                                                    \\
                                     & road condition                           & \multicolumn{2}{l}{normal /busy/congested}                                                                                                             \\ 
\hline
\multirow{13}{*}{Object Description} & number of lanes                          & \multicolumn{2}{l}{integer}                                                                                                                            \\ 
\cline{2-4}
                                     & ego lane number                          & \multicolumn{2}{l}{integer}                                                                                                                            \\
                                     & ego lane condiiton                       & \multicolumn{2}{l}{normal/congested}                                                                                                                   \\
                                     & ego lane type                            & \multicolumn{2}{l}{dashed lane/solid lane/dashed-solid lane/solid-dashed lane}                                                                         \\
                                     & ego actions                               & \multicolumn{2}{l}{forward/stop/go ahead to the left/go ahead to the right}                                                                            \\ 
\cline{2-4}
                                     & critical object description              & \multicolumn{2}{l}{like white car, red van, electric car, pedestrian and so on}                                                                        \\
                                     & critical object lane number              & \multicolumn{2}{l}{integer}                                                                                                                            \\
                                     & critical object relative position        & \multicolumn{2}{l}{left/right/front/left front/right front}                                                                                            \\
                                     & critical object distance                 & \multicolumn{2}{l}{normal/relatively close/dangerous}  \\ 
\cline{2-4}
                                     & \multirow{4}{*}{critical object actions} & general actions & forward/slow down/stop/go ahead to the left/go ahead to the right                                                                    \\
                                     &                                          & vehicle actions & overtake/pull into ego lane to the left/ pull into ego lane to the right                                                             \\
                                     &                                          & vehicle signals & no turn signal/brake                                                                                                                 \\
                                     &                                          & other actions   & crossing the street from left to right/ crossing the street from right to left                                                       \\
\hline
\end{tabular}}
\label{tab:data-labeling-details}
\end{table}